\documentclass[conference]{IEEEtran}

\usepackage{cite}
\usepackage{amsmath,amssymb,amsfonts}
\usepackage{algorithmic}
\usepackage{graphicx}
\usepackage{textcomp}
\usepackage{xcolor}

\usepackage{booktabs}
\usepackage{hyperref}
\usepackage{url}   
\usepackage{soul}
\usepackage{amsmath}
\usepackage{bm}
\usepackage{amssymb}
\usepackage{subcaption}
\usepackage{placeins}
\begin{document}

\newcommand\blfootnote[1]{%
  \begingroup
  \renewcommand\thefootnote{}\footnote{#1}%
  \addtocounter{footnote}{-1}%
  \endgroup
}
\renewcommand*{\thefootnote}{\alph{footnote}}

\title{Cost-Sensitive Diagnosis and Learning Leveraging Public Health Data}
\author{\IEEEauthorblockN{Mohammad Kachuee, Kimmo Karkkainen, Orpaz Goldstein, Davina Zamanzadeh, and Majid Sarrafzadeh}
\IEEEauthorblockA{\textit{Department of Computer Science} \\
\textit{University of California, Los Angeles (UCLA)}\\
}}

\maketitle

\begin{abstract}

Traditionally, machine learning algorithms rely on the assumption that all features of a given dataset are available for free. However, there are many concerns such as monetary data collection costs, patient discomfort in medical procedures, and privacy impacts of data collection that require careful consideration in any real-world health analytics system. An efficient solution would only acquire a subset of features based on the value it provides while considering acquisition costs. Moreover, datasets that provide feature costs are very limited, especially in healthcare. In this paper, we provide a health dataset as well as a method for assigning feature costs based on the total level of inconvenience asking for each feature entails. Furthermore, based on the suggested dataset, we provide a comparison of recent and state-of-the-art approaches to cost-sensitive feature acquisition and learning. Specifically, we analyze the performance of major sensitivity-based and reinforcement learning based methods in the literature on three different problems in the health domain, including diabetes, heart disease, and hypertension classification.
\end{abstract}



\section{Introduction}
\blfootnote{The related source code and data for this paper is available at \url{https://github.com/mkachuee/Opportunistic}}
Traditional machine learning is focused on modeling dynamics of a dataset consisting of features that are freely available.
However, in many real-world problems, especially in the health domain, having access to the value of each feature entails a certain cost which requires careful consideration. This notion of cost is general and may include the actual monetary cost, patient discomfort, privacy impacts, and so forth \cite{krishnapuram2011cost,pattuk2015privacy,hatamizadeh2018automatic,ghasemzadeh2015power}.
Careful consideration of these costs and devising algorithms and methods that consider this notion can be crucially important in health settings as it can reduce the data collection costs and increase the human subject compliance. In a cost-sensitive learning scenario, information is being acquired based on optimizing the balance between the predictive value it provides and the cost entailed by the acquisition. This is in contrast to the traditional approach frequently being used in health care studies which requires iterative expert hypothesis, pilot data collections, and analysis.


Perhaps the most naive approach to the problem of feature acquisition is feature selection. Feature selection methods select a fixed subset of available features and limit the analysis to that subset\cite{melville2005economical,ghasemzadeh2015power,efron2004least}. For instance, Lasso\cite{tibshirani1996regression}, an L1 regularized linear model, enforces a level of sparsity for model weights which effectively limit the features to be used. Note that these methods neglect the available context at the prediction time which leads to solutions that are not optimal.

More recently, probabilistic modeling and inference approaches in the context of decision making in Bayesian networks have been suggested as more theoretically convincing approaches. Chen \textit{et al.}\cite{chen2015value} introduced same decision probability (SDP) to measure the probability of change in decisions given new evidence as a measure of feature value and prediction confidence. While this method is theoretically plausible, SDP is computationally expensive and is mostly applicable to binary features and Bayesian networks\cite{chen2014algorithms}. This renders SDP and similar probabilistic approaches that rely on explicit distribution modeling less applicable to healthcare problems where the number of features is usually large and many features are real valued. 

An alternative approach which is frequently used in the literature is making cost-sensitive predictors based on cascade or tree classifiers\cite{karayev2012timely,chen2012classifier,xu2012greedy,xu2014classifier}. Perhaps, the classification cascade by Viola~\textit{et al.}\cite{viola2004robust} is the most famous example of these methods, reducing the computational cost via early rejection. Nan~\textit{et al.}\cite{nan2017adaptive} suggested a gating mechanism between low prediction cost and high prediction cost models. Decision trees are naturally designed to make decisions via prioritizing features that provide the highest information gain. For healthcare analytics, these methods are promising in terms of model interpretability; however, due to a fixed structure, their feature query decisions are not truly instance specific. In addition, these methods are limited by the shortcomings of decision trees such as modeling issues with a large number of features and greedy decisions during the tree generation.

Recently, sensitivity analysis of predictor models was suggested as a measure of feature importance \cite{early2016test,kachuee2018dynamic}. In this context, sensitivity is a feature value measure representing the influence of each feature on the predicted outcome by a predictor model. The main advantage of these methods is the fact that they are usually compatible with available prediction algorithms and require less implementation effort.

Alternatively, reinforcement learning solutions were suggested that formulate the feature acquisition and prediction process as a Markov decision process \cite{peng2018refuel,janisch2017classification,shim2017pay,kachuee2019opportunistic}. While reinforcement learning approaches are more flexible and powerful compared to other counterparts, training these models is usually complicated, and the definition of the reward function plays a crucial role in the final performance.

While there has been good progress in the development of these algorithms and methods, there has been little work done on the evaluation and application of them on real-world data in general, and health data in particular. The main reason behind this is that the currently available datasets rarely provide feature costs. Consequently, arbitrary or synthesized cost assignments are frequently being used in the literature which prevents the evaluation of these methods in actual use cases such as disease diagnosis.

In this paper, we provide a study of cost-sensitive learning for smart health scenarios. Specifically, we provide a framework for mining datasets from public health records released by CDC. It consists of demographics, examination, questionnaire, and laboratory data for about 100,000 individuals. We propose a methodological way for real cost assignment based on a survey conducted using Amazon Mechanical Turk. Furthermore, we present a comparison of major cost-sensitive learning methods on this dataset and across various problems.

\section{Methodology}
\subsection{Data Source}
We use the National Health and Nutrition Examination Survey (NHANES) \cite{nhanes} between 1999 to 2016 as our data source. NHANES is an ongoing survey which is designed to assess the well-being of adults and children in the United States. Each year, health and nutrition data is collected from few thousand individuals and consisting of demographics, questionnaire, examination, and laboratory data. Not all data is collected from each individual (e.g., certain blood tests are not used for young children) and there is a slight variation between the information being collected in each year (e.g., the prevalence of disease change over time causing changes on the data collection focus). For more information about NHANES please refer to the documentation\cite{nhanes}$^,$\footnote{\url{https://www.cdc.gov/nchs/nhanes}}.

\subsection{Data Preparation}
\label{sec:Data Preparation}
We developed a general data processing pipeline which can be used for different tasks and settings. The data preparation starts with loading raw data files associated with each variable in the dataset containing values of that variable for each subject. Here, a variable can be answer to a demographics question, a certain factor in a blood test, result of a certain examination, etc. Please note that we merge columns and rows based on variable and subject identifiers so that, logically, all variables appear as columns and individuals appears as a row. For a certain task, any available variable could be defined as a feature or a target, depending on the task.

The dataset consists of $9385$ unique variables of different types including categorical, real-valued, multiple choice, etc. Accordingly, we use different preprocessing functions such as statistical normalization for real-valued variables and one-hot encoding for categorical variables to prepare each variable for further analysis.
Also, it should be noted that, for each individual, only a subset of these variables is available and the rest are missing.

To define each certain task (e.g. diabetes classification), a target should be defined as a function of available variables (e.g.,  blood glucose level categories). 
Two methods are suggested to determine variables to be used as input features and appropriate preprocessing functions : $(i)$ explicitly defining a list of variables and their corresponding preprocessing functions $(ii)$ automatically selecting relevant features by searching over the space of variables and automatically deciding on appropriate preprocessing functions. Regarding the second method, one can limit the number of features being eventually used by setting a threshold on the mutual information between the target and each feature as well as a threshold on the percentage of available (not missing) features for each selected variable. Applying this thresholds limits the feature set to features that are informative for the task and are available for a certain percentage of dataset samples.

\begin{figure}[t]
\centering
  \includegraphics[width=0.99\linewidth]{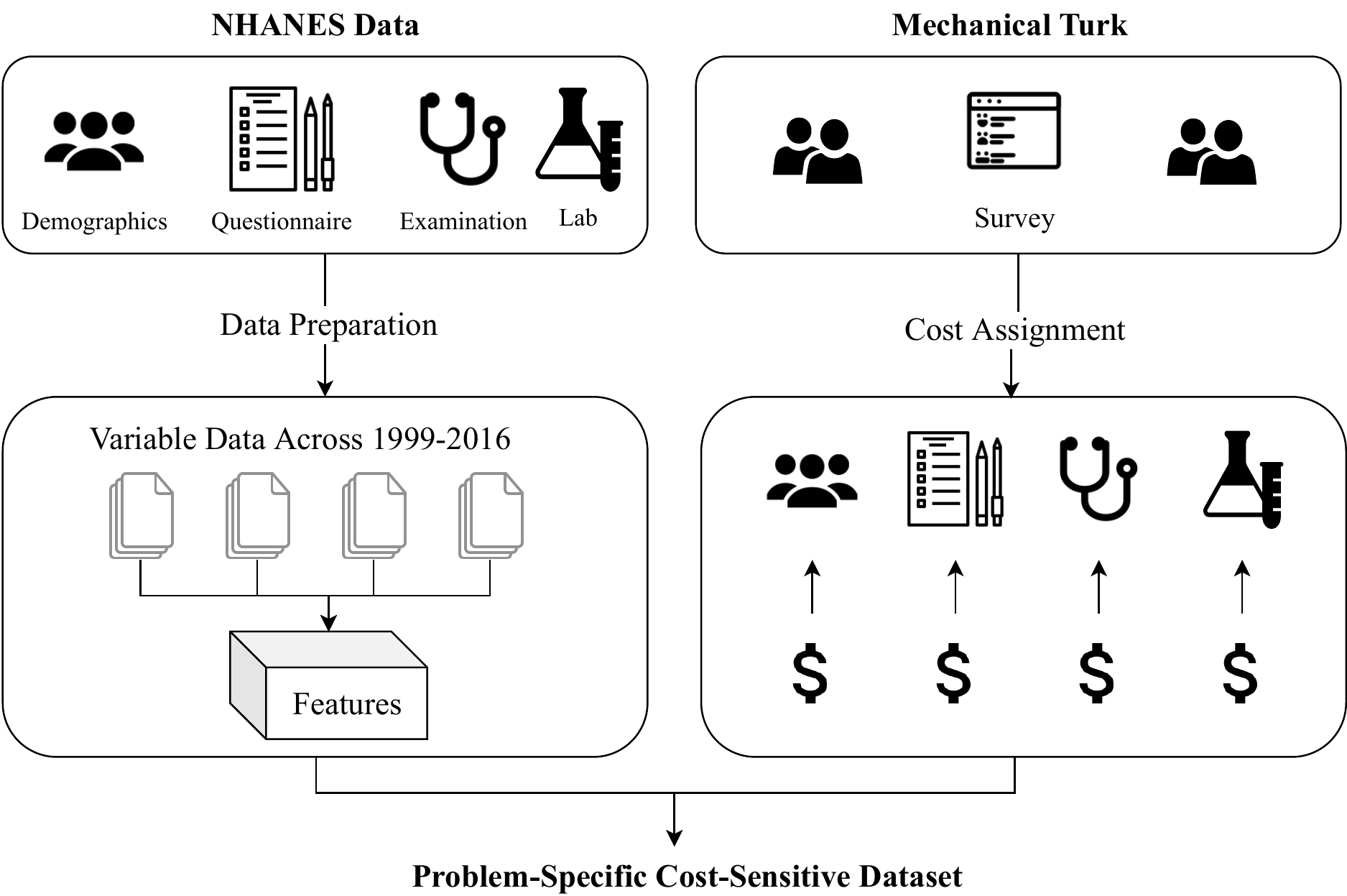}
  \caption{Visualization of the proposed preprocessing and cost assignment pipeline.}
  \label{fig:flowchart}
\end{figure}

\subsection{{Cost Assignment}}

\begin{table*}[h!]%
\vspace{0.3in}
\caption{The questionnaire used in this study.}
\label{tab:survey}
\renewcommand{\arraystretch}{1.3}
\begin{center}
\resizebox{2.0\columnwidth}{!}{
\begin{tabular}{lll}
\toprule
\textbf{No.} & \textbf{Question} & \textbf{Answer}\\
\midrule
1 & Convenience of answering general demographics related questions (e.g., age, gender, race, etc.) & 1\dots10 \\
2 & Convenience of answering general behavioral/life-style related questions (e.g., smoking habits, sleeping habits, alcohol consumption, drug usage etc.) & 1\dots10 \\
3 & Convenience of getting typical examinations such as weight, height, or blood pressure measurement & 1\dots10 \\
4 & Convenience of taking a blood or urine test at a lab & 1\dots10 \\
\bottomrule
\end{tabular}
}
\end{center}
\end{table*}%

\begin{figure*}[h]
    \centering
    \begin{subfigure}[b]{0.23\textwidth}
        \centering
        \includegraphics[width=\textwidth]{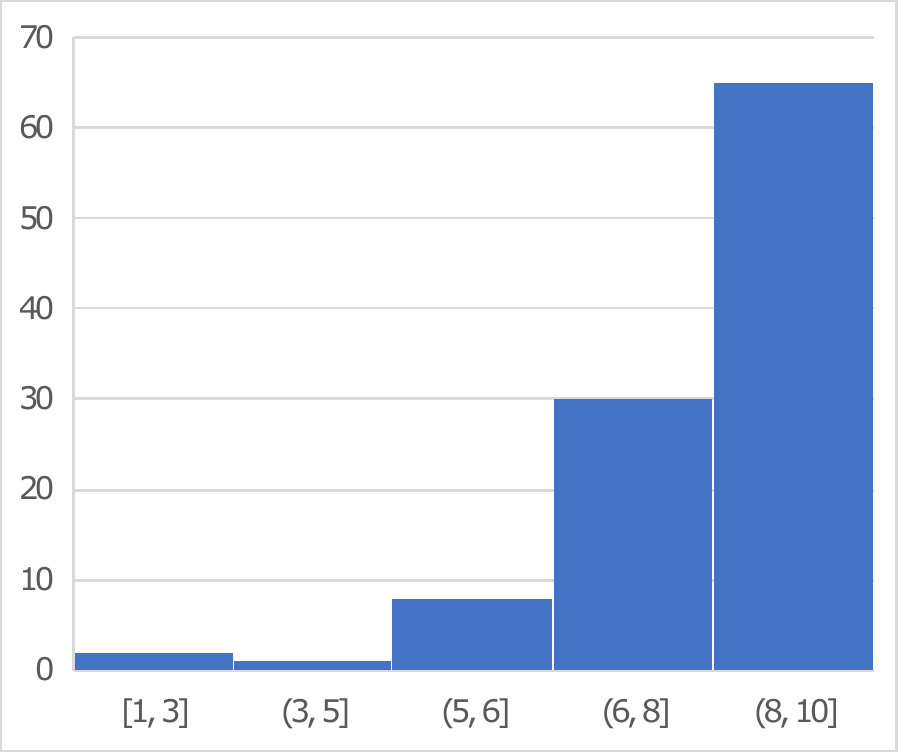}
        \caption{}
    \end{subfigure}%
    ~ 
    \begin{subfigure}[b]{0.23\textwidth}
        \centering
        \includegraphics[width=\textwidth]{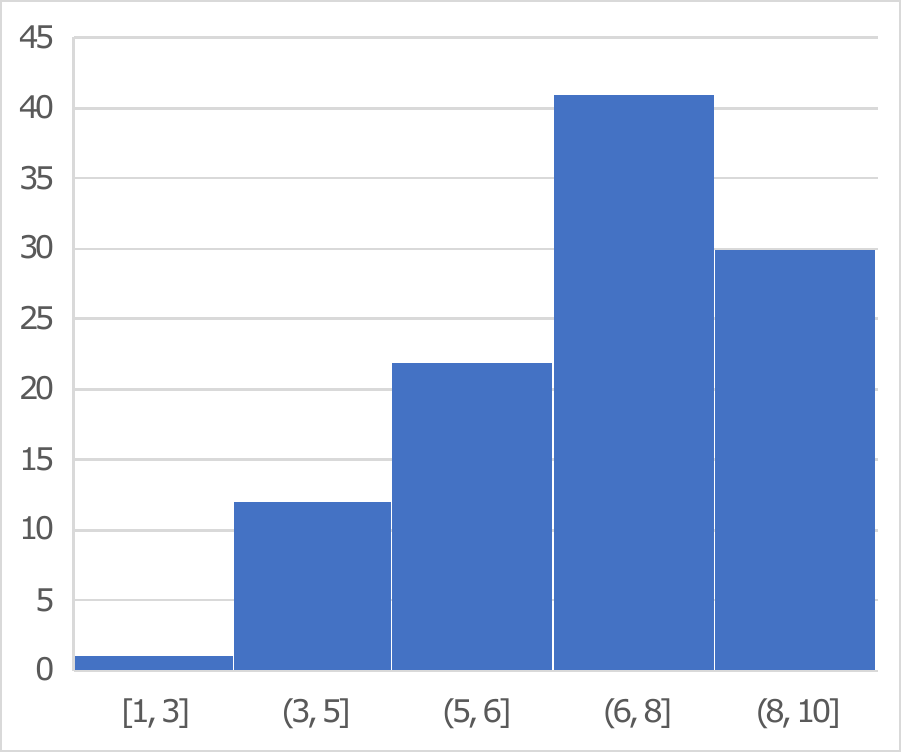}
        \caption{}
    \end{subfigure}
    ~
    \begin{subfigure}[b]{0.23\textwidth}
        \centering
        \includegraphics[width=\textwidth]{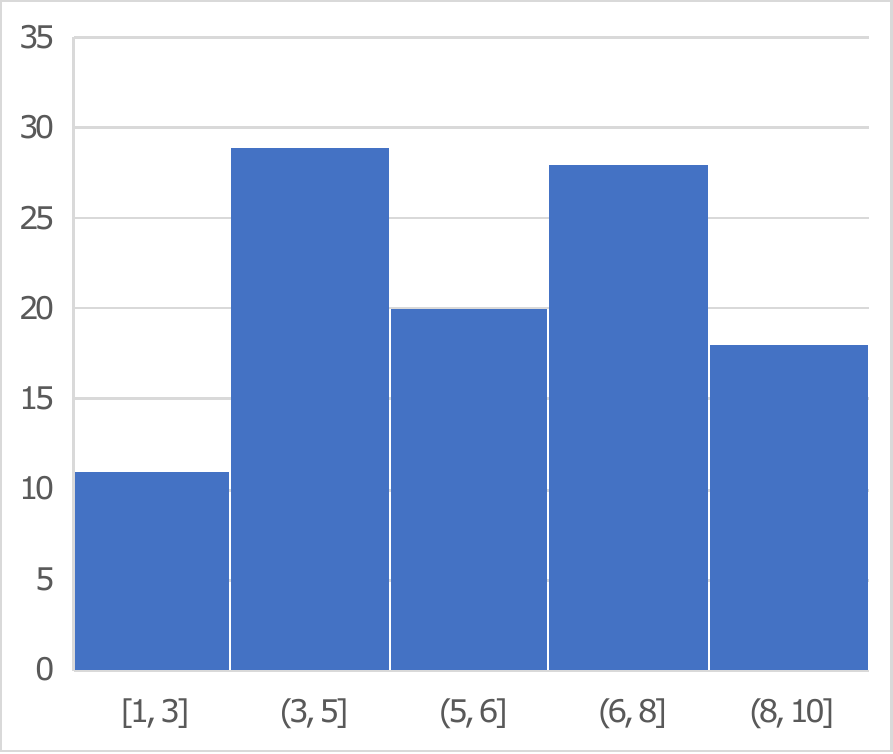}
        \caption{}
    \end{subfigure}
    ~
    \begin{subfigure}[b]{0.23\textwidth}
        \centering
        \includegraphics[width=\textwidth]{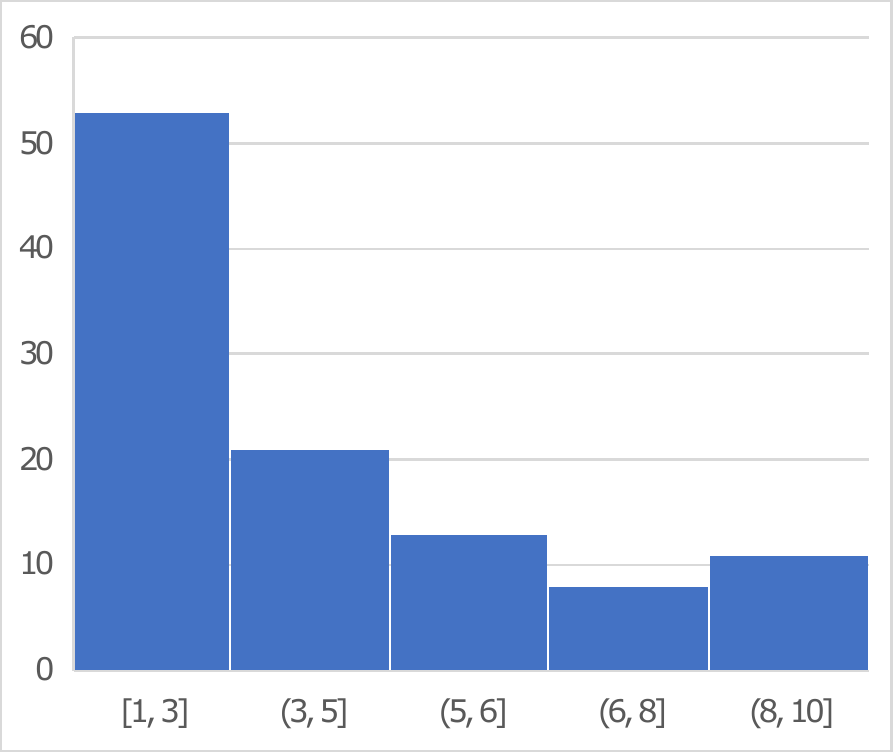}
        \caption{}
        \label{fig:mnist_vis_multires}
    \end{subfigure}
    \caption{Distribution of answers for the level of convenience collecting information about: (a) demographics, (b) behavioral/life-style, (c) medical examinations, and (d) lab tests.}
    \label{fig:hist_cost}
\end{figure*}

In order to assign costs corresponding to each feature, we conducted a survey to collect the level of overall inconvenience that asking for each feature would cause to subjects. Specifically, we collected survey data from $108$ individuals using the Amazon Mechanical Turk framework. Since the data is collected from individuals in the United States, we limited our survey to the same population. Table~\ref{tab:survey} presents the questionnaire used in this study. Before starting the questionnaire, we asked the turkers to pay attention to the following instructions:
\begin{itemize}
    \item Please rate each question in terms of the total inconvenience they will cause you (including time burden, financial cost, discomfort, etc.).
    \item Assume that each item will provide you with useful health information; however, there is no urgency to do any of these.
    \item The scale is $1$ to $10$, $10$ being the most convenient. Rate each item based on the relative level of inconvenience.
    \item After completing the sheet, please review and adjust, if necessary.
\end{itemize}

The median of survey results for each question is used as a level of convenience for each question. In order to convert these values to cost values (i.e., the higher the more expensive), we subtract each convenience value from $11$ and consider the resulting value as the cost of acquiring features of that category. Accordingly, the final cost for feature categories corresponding to questions 1 to 4 of Table~\ref{tab:survey} is determined to be $2$, $4$, $5$ and $9$, respectively. See Figure~\ref{fig:flowchart} for an overview of the data preparation methodology and Figure~\ref{fig:hist_cost} for the distribution of answers.


\subsection{{Problem Definition}}
In this paper, we consider the general scenario of supervised classification using a set of features $\bm{x}_i \in \mathbb{R}^d$ and ground truth target labels $\tilde{y_i}$. However, initially, the values of features are not available and there is a cost for acquiring each feature ($\bm{c}_j; 1 \le j \le d$).
Consequently, for sample $i$, at each time step $t$, we only have access to a partial realization of the feature vector denoted by $\bm{x}_i^t$ consisting of features that are acquired until $t$. 
There might be a maximum budget ($B$) or a user-defined termination condition (e.g., prediction confidence) which limits the features being acquired eventually.

More formally, we define a mask vector $\bm{k}_i^t \in \{0,1\}^d$ where each element of $\bm{k}$ indicates if the corresponding feature is available in $\bm{x}_i^t$. Using this notation, the total feature acquisition cost at each time step can be represented as
\begin{equation}
    C^t_{total, i} = (\bm{k}_i^t - \bm{k}_i^0)^T \bm{c} ~~ .
\end{equation}
Intuitively, it measures the total budget spent from the initial state to the current state at time $t$. Furthermore, we define the feature query operator (q) as 
\begin{equation}
    \bm{x}_i^{t+1} = q(\bm{x}_i^t, j), \text{ where } \bm{k}_{i,j}^{t+1} - \bm{k}_{i,j}^t = 1 ~~ .
\end{equation}

Note that acquiring a feature at time t results in the transition to a new state, $t+1$, having access to the value of that feature. In this setting, the objective of a cost-sensitive feature acquisition algorithm is to balance the accuracy versus cost trade-off via efficiently acquiring as many features as necessary at test-time.

\subsection{{Sensitivity-Based Approach}}
Sensitivity-based approaches use trained classifier models and select features that have the most influence on the model predictions. This influence can be used as a utility function which measures the importance of having access to each feature value.

Early \textit{et al.} \cite{early2016test} suggested an exhaustive measurement of expected sensitivity for each feature:
\begin{equation}
\label{eq:exh}
    \mathbb{E}[U(x^t,j)] = \int p(x_j=r|x^t) U(x^t,x_j=r) dr , 
\end{equation}
where $U(x^t,j)$ is the expected utility of acquiring feature $j$ given the feature vector at $t$, and with the abuse of notation, $U(x^t,x_j=r)$ is the utility of that feature assuming its value after acquisition would be equal to $r$.
It should be noted that as this method requires modeling joint probability distributions as well as integration over feature values, it is not scalable to datasets consisting of many features.

In an earlier work\cite{kachuee2018dynamic}, the authors suggested an approximation of \eqref{eq:exh} using a binary representation layer in denoising autoencoder architectures:
    \begin{equation}
    j^{t}_{sel} = \underset{j \in \{1 \dots d\}|k_j^t=0}{argmax} \frac{\sum_{x_{j}\in RS} |\frac{\partial h(\bm{x}^t)}{\partial x_{j}}|\: p(x_{j}|\bm{x}^t;h^t)}{c_j^t} \;, 
    \label{eq:opt_summation}
    \end{equation}
    where $RS = \{2^{-l}, \dots, 2^{-2}, 2^{-1}, 1\}$ and $h(x^t)$ is the classification function. It is worth noting that this approximation leads to much faster computation based on a single forward and backward evaluation of the network. Specifically, approximating $p(x_{j}|\bm{x}^t;h^t)$ terms using a denoising autoencoder with a binary layer and approximating $|\frac{\partial h(\bm{x}^t)}{\partial x_{j}}|$ terms via a backpropagation from outputs to the binary input layer.  


\subsection{{Reinforcement Learning Approach}}
The cost-sensitive feature acquisition problem can be formulated as a reinforcement learning problem. In this setting, each state would be the features that are acquired at each point. Additionally, each action would be to pay for a certain feature and to acquire its value, transitioning to a new state. Here, the objective would be to learn a policy function that results in an efficient feature acquisition. One possible reward function which is frequently used in the literature \cite{janisch2017classification,shim2017pay} is:
\begin{equation}
\label{eq:reward_rl} 
    r(x_i^t, a) = \begin{cases}
        -\bm{c}_j & \text{$a$ is acquiring feature $j$}\\
        0 & \text{$a$ is making a correct prediction}\\
        -\lambda & \text{$a$ is making an incorrect prediction}\\
    \end{cases} ,
\end{equation}
where $\lambda$ is a hyperparameter controlling the acquisition cost and prediction accuracy trade-off. In this equation, $-\lambda$ is to penalise incorrect predictions and $-\bm{c}_j$ is to penalise acquiring each feature according to its cost. Therefore, using higher $\lambda$ values increases the prediction accuracy while increasing the total cost expenditures.

Alternatively, the variations of model certainty weighted by feature costs can be used to define a reward function \cite{kachuee2019opportunistic}: 
\begin{equation}
\label{eq:reward_ol} 
    r_{i,j}^t =  \frac{\Vert Cert(\bm{x}_i^t) - Cert(q(\bm{x}_i^t, j))\Vert}{\bm{c}_j},
\end{equation}
where $Cert(x)$ represents the prediction certainty\cite{gal2016dropout} using a feature vector x. Intuitively, it measures the value of each feature based on the amount which having access to that feature would contribute to making more confident predictions. This method is shown to offer the state of the art results for cost-sensitive feature acquisition at test-time\cite{kachuee2019opportunistic}. Furthermore, it is highly scalable and applicable to online stream processing. This can be particularly useful in clinical setups in which prompt decisions are vitally important. For instance, prescribing certain tests and making fast and yet reasonably accurate diagnosis can be life saving.

\section{Experiments}


In order to evaluate and provide baselines for the proposed dataset, in this section, we define specific cost-sensitive classification tasks and present comparison results for several recent cost-sensitive learning methods. Specifically, we compare: $i)$ a method based on reinforcement learning where a hyper-parameter is balancing the cost vs. accuracy trade-off \cite{janisch2017classification}, $ii)$ Opportunistic Learning (OL) \cite{kachuee2019opportunistic} a method based on deep Q-learning with variations of model uncertainty as the reward function, $iii)$ a method based on exhaustive measurements of the sensitivity \cite{early2016test}, and $iv)$ a method based on approximation of sensitivities using denoising autoencoders (FACT) \cite{kachuee2018dynamic}. Table~\ref{tab:experiments} presents a summary of dataset tasks we defined including the number of instances, features, and classes as well as baseline classification accuracies for each task.

\begin{table}[t]%
\caption{The summary of datasets and experimental settings.}
\label{tab:experiments}
\renewcommand{\arraystretch}{1.3}
\begin{center}
\begin{tabular}{cccccc}
\toprule

\textbf{Task} & \textbf{Instances} & \textbf{Features} & \textbf{Classes} & \textbf{Baseline Accuracy} \\
\hline
Diabetes & 92062 & 45 & 3 & 84.2\%\\ 
Heart Disease & 49509 & 97 & 2 & 79.7\%\\
Hypertension & 22270 & 31 & 2 & 81.9\%\\  
\bottomrule
\end{tabular}
\end{center}
\end{table}%

We use PyTorch computational library\cite{paszke2017automatic} to train and evaluate multi-layer neural network architectures. Throughout experiments, adaptive momentum (Adam) is used as the optimization algorithm\cite{kingma2014adam}. Each experiment took between a few hours to a day on a GPU server. It is worth noting that we normalize dataset features to have zero mean and unit variance prior to our experiments. This normalization permits using the value of zero for missing features during the prediction to act as mean imputation. Regarding the number of layers and hidden neurons, we used a similar number of trainable parameters for OL\cite{kachuee2019opportunistic} and RL-Based\cite{janisch2017classification}, while due to the inherent differences, we had to use different architectures for FACT\cite{kachuee2018dynamic} and Exhustive\cite{early2016test}. Nonetheless, for each classification task, the compared models reach a similar baseline accuracy (i.e., average accuracy after acquiring all the features). In the following, we provide a brief explanation of each task. For more details about specific features and setups please refer to the Git page\footnote{\url{https://github.com/mkachuee/Opportunistic}}.

\begin{figure}[t]
\centering
  \includegraphics[width=0.9\linewidth]{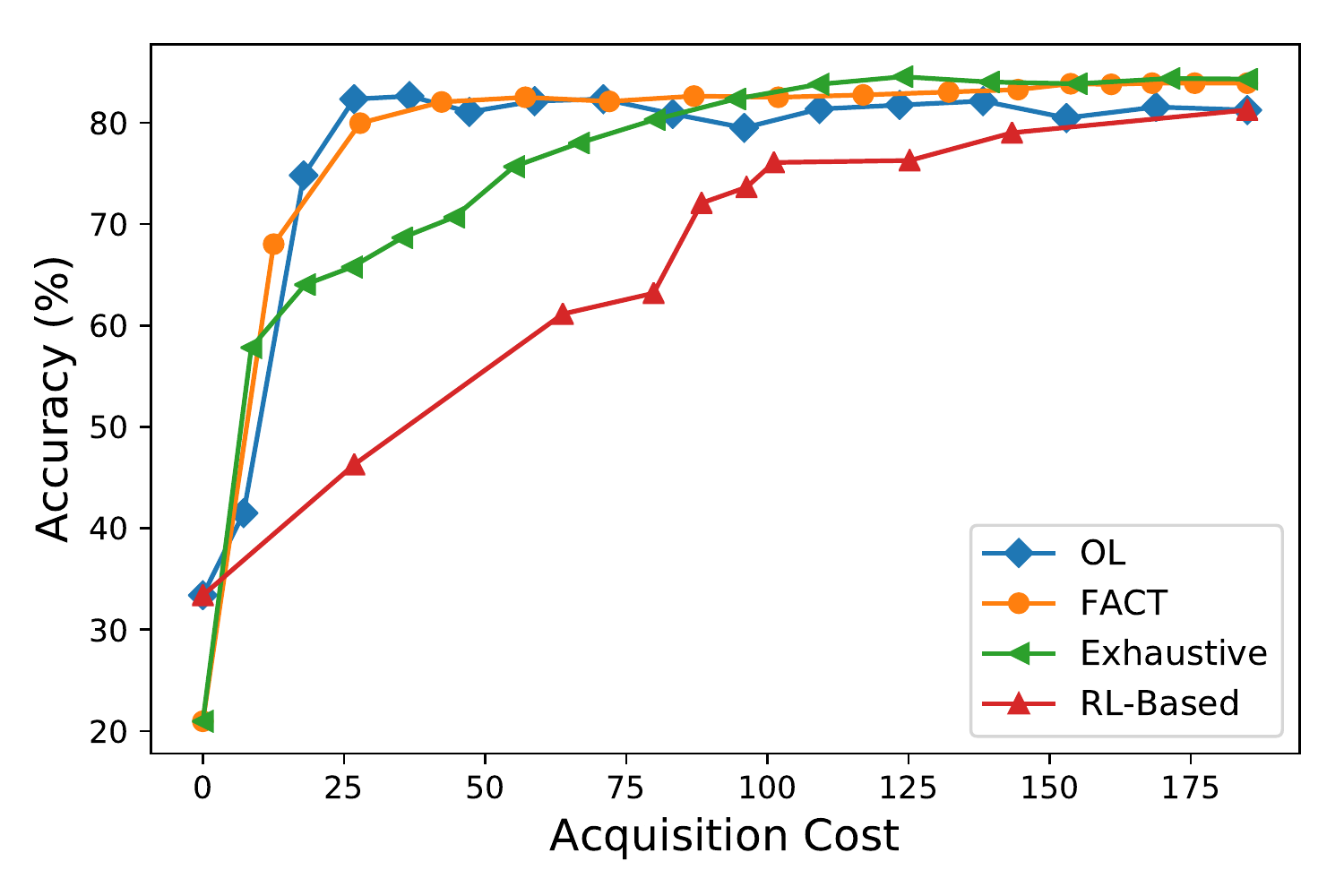}
  \caption{Accuracy versus cost curve for the diabetes classification task comparing OL \cite{kachuee2019opportunistic}, FACT \cite{kachuee2018dynamic}, Exhaustive \cite{early2016test}, and RL-Based \cite{janisch2017classification} methods.}
  \label{fig:diabetes}
\end{figure}
\begin{figure}[t]
\centering
  \includegraphics[width=0.9\linewidth]{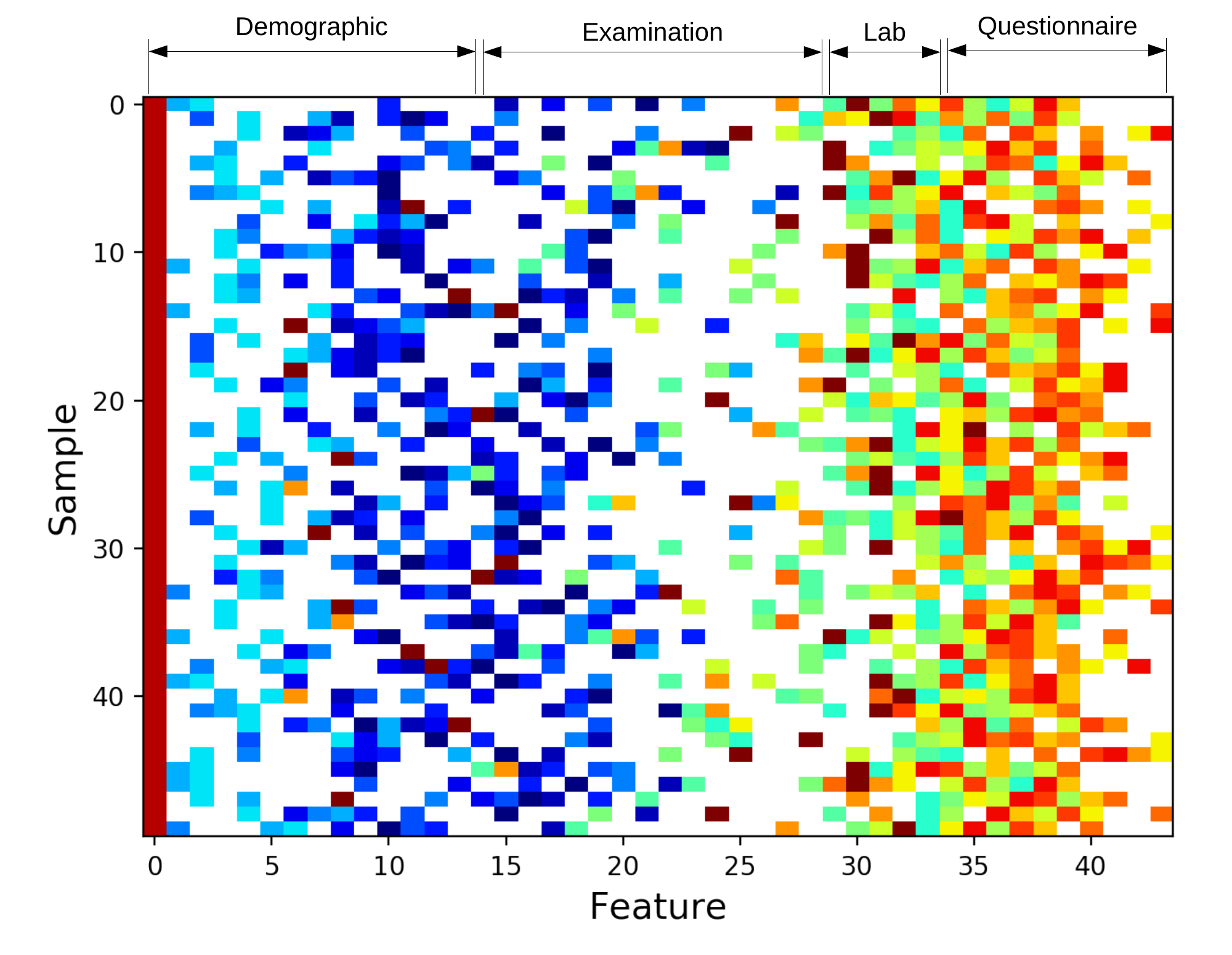}
  \caption{Visualization of feature acquisition orders using OL \cite{kachuee2019opportunistic} method on the diabetes dataset. Warmer colors indicate more priority.}
  \label{fig:feature_order}
\end{figure}

We defined diabetes as the classification objective to predict the blood glucose categories that fall into the following three categories: normal (blood sugar less than 100 mg/dL), prediabetes (blood sugar between 100-125 mg/dL), and diabetes (blood sugar more than 125 mg/dL). We specifically defined and used 45 relevant features from demographics, questionnaire, examinations, and lab tests. Figure~\ref{fig:diabetes} presents a comparison of results based on this task. As it can be seen from this figure, FACT and OL achieve superior results compared to other work. Figure~\ref{fig:feature_order} provides a visualization of feature acquisition order of 50 randomly selected samples for demographic, examination, laboratory, and questionnaire feature categories. In this figure, features that are acquired by OL with more priority are indicated with warmer colors. As it can be inferred from this figure, questionnaire information are usually being acquired with more priority as they provide valuable information at a reasonable cost.

\begin{figure}[t]
\centering
  \includegraphics[width=0.99\linewidth]{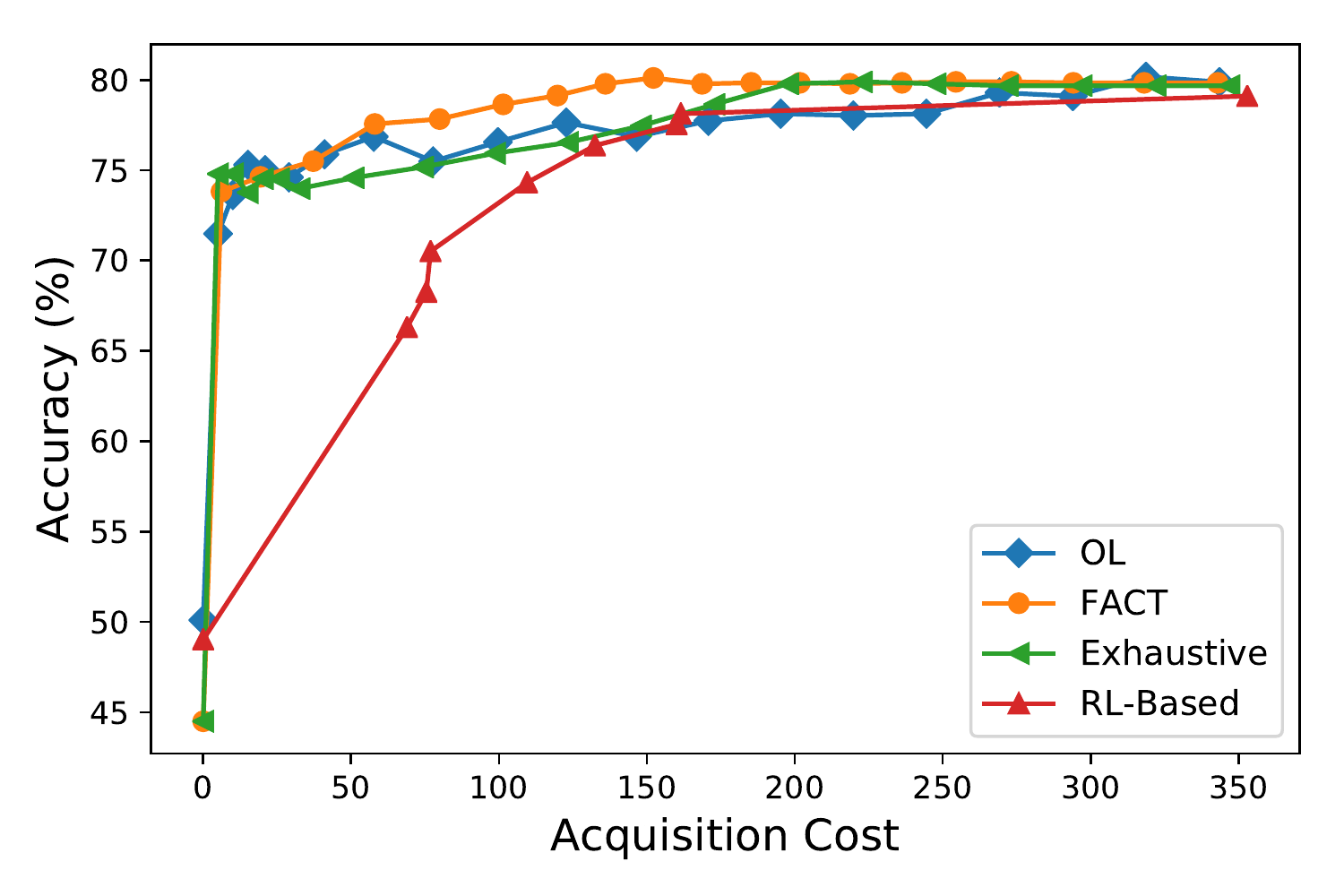}
  \caption{Accuracy versus cost curve for the heart disease classification task comparing OL \cite{kachuee2019opportunistic}, FACT \cite{kachuee2018dynamic}, Exhaustive \cite{early2016test}, and RL-Based \cite{janisch2017classification} methods.}
  \label{fig:heart}
\end{figure}

\begin{figure*}[t]
\centering
  \includegraphics[width=0.7\linewidth]{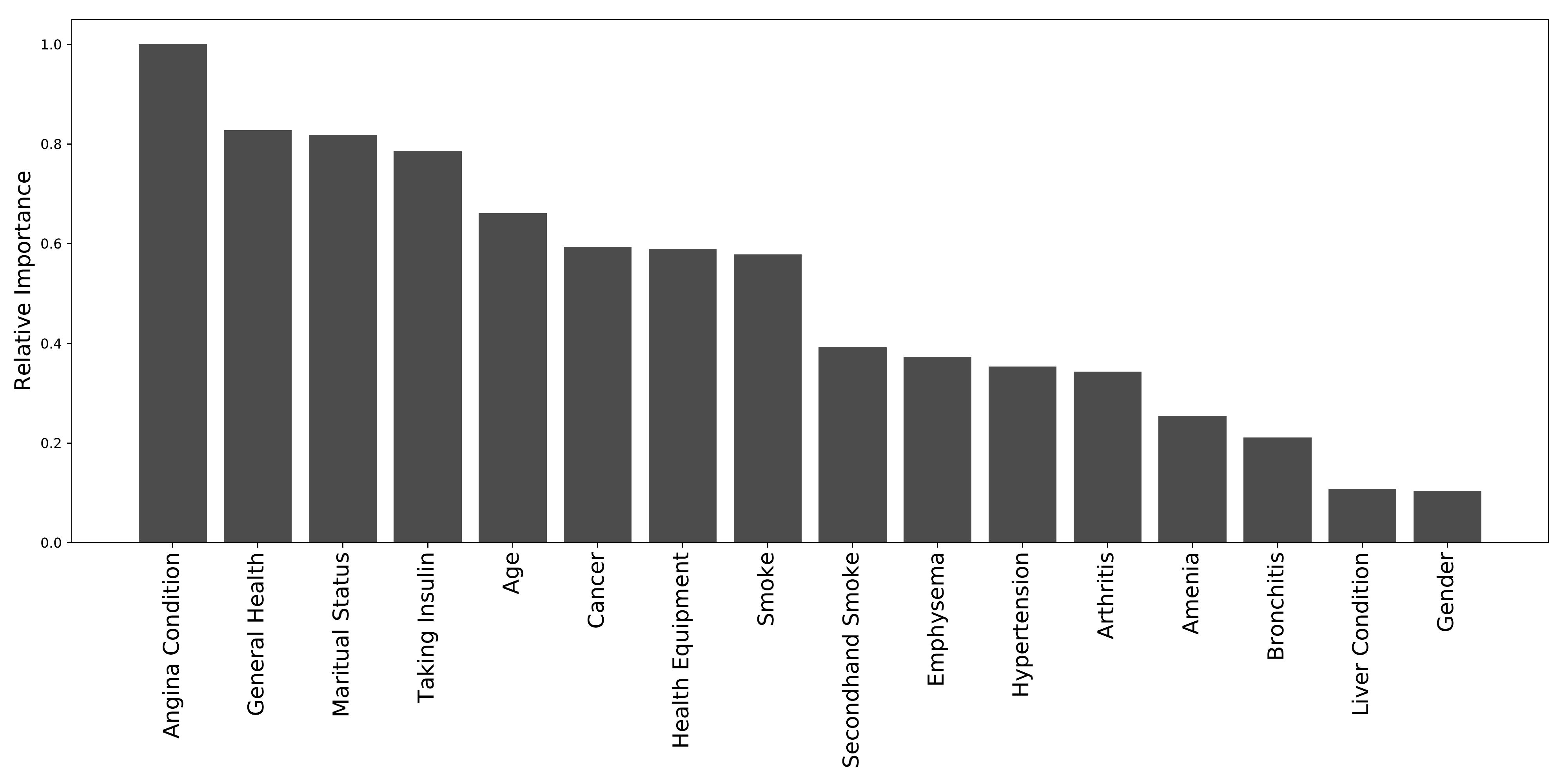}
  \caption{The relative importance for the 16 most relevant features used in the heart disease classification task.}
  \label{fig:heart_features}
\end{figure*}

As another task, we consider heart disease classification which is defined as binary classification task. Here, positive samples are individuals that reported any heart disease related issue in their history (e.g., heart attack, heart failure, etc.). For this task, we used the automated feature selection method as explained in the Data Preparation section resulting in 97 features. Figure~\ref{fig:heart_features} shows the relative importance for the 16 most relevant features used in the heart disease classification task. Here, the feature importance is measured based on the relative magnitude of weights for a logistic classifier trained on this dataset. As it can be seen from this figure, the suggested automated variable selection used for this task selects features that are intuitively relevant to heart conditions. Furthermore, Figure~\ref{fig:heart} presents the comparison of results for this task. It can be inferred that this task is relatively simple and most approaches were able to achieve a reasonable performance.

\begin{figure}[t]
\centering
  \includegraphics[width=0.99\linewidth]{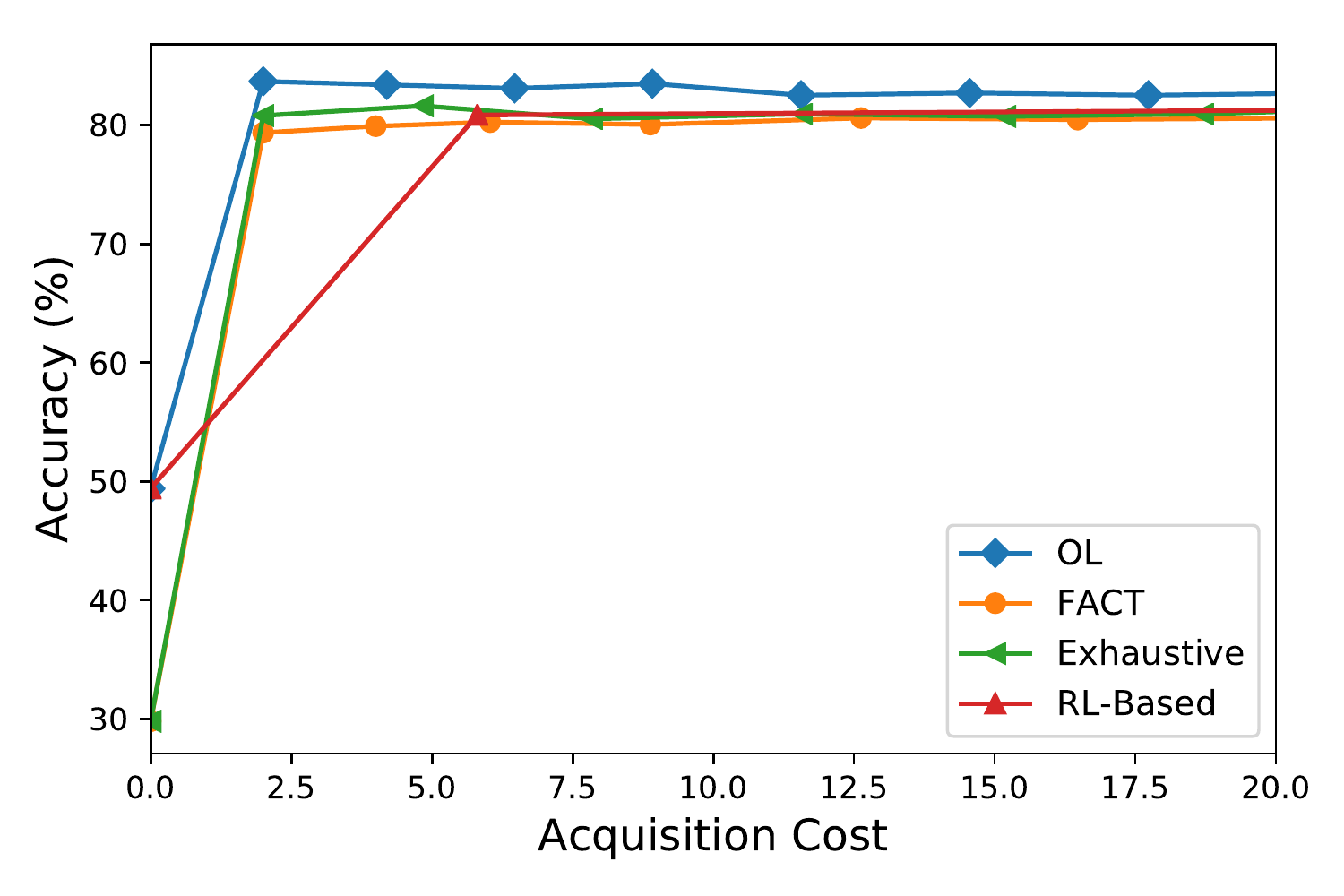}
  \caption{Accuracy versus cost curve for the hypertension classification task comparing OL \cite{kachuee2019opportunistic}, FACT \cite{kachuee2018dynamic}, Exhaustive \cite{early2016test}, and RL-Based \cite{janisch2017classification} methods.}
  \label{fig:hypertension}
\end{figure}

At last, we consider the problem of predicting the existence of hypertension condition in individuals. Specifically, we consider subjects with systolic blood pressure of more than 140~mmHg as positive (hypertensive) class. Figure~\ref{fig:hypertension} shows the performance of different methods on this dataset. It can be inferred from this figure that this task is relatively easy and all methods were able to achieve a reasonable performance. The only exception is the RL-based method which we were not able to find  hyper-parameters resulting in cost values in the range of 0 and 5.

\section{Discussion}

Classical approaches to machine learning sought to improve the efficiency and accuracy of prediction but often failed to account for the costs associated with the collection of data and expert labels; the acquisition of some features might incur more costs (monetary and non-monetary) than others. This shortcoming is particularly limiting in the health informatics, where accurate classification often requires an invasive level of information querying. Furthermore, in domains such as medical diagnosis, appropriate data should be collected based on scientific hypothesis, and ground-truth labels may only be provided by highly trained domain experts. Additionally, in many studies, informative features are not scientifically predetermined, and usually, there are many information sources that can be considered as hypothetical relevant features which including all of them is not practical. This is one of the reasons healthcare data remains underutilized. With the explosion of data that the world of IoT brings to the healthcare domain, solutions that are able to efficiently take the unique requirements of smart-health, and scale effectively to the amount of data that is available, are needed. Developing intelligent aggregation methods that leverage available data, enables us to collect the right information at the right time.

In order to address these issues, cost-sensitive and context-aware learning methods were suggested that consider different aspects of a real-world learning-based system. Specifically, addressing all components needed in such a system; including feature acquisition, labeling, model training, and prediction at test-time, each trying to achieve the goal of making accurate predictions efficiently. In this paradigm, information is acquired incrementally based on the value it provides and the cost that should be paid for acquiring it. For instance, in the medical diagnosis use case, while asking for an MRI scan might be more informative than a simple blood test, asking for the blood test might be a better option to start with, based on the information gain per unit of cost for a blood test. This can potentially result in huge monetary and non-monetary savings. Finally, in this paradigm, data collection and training are coupled with each other which results in collecting data as much as required while making accurate predictions. This is in contrast to the traditional machine learning setup which collecting a reasonable size of data is assumed to be happening before conducting any analysis.

In this paper, we explored dynamic and context-aware information acquisition techniques to collect the right piece of information at the right time. The proposed solution, in healthcare settings, enhances the human subject compliance and experience by reducing the cost and inconvenience of medical tests and the data collection. Authors have extensively studied these techniques under the name of "Dynamic Cost-Aware Feature Acquisition"\cite{kachuee2018dynamic} and "Opportunistic Learning"\cite{kachuee2019opportunistic}. In this framework, each information piece has a cost which can be predetermined by the user or the expert. In this study, we presented a comparison between the applicability of these methods as well as related work in the literature to health domain problems. Additionally, the scarcity of datasets in healthcare that provide feature costs limited the application of cost-sensitive methods in this domain. In this paper, we suggested a methodology for creating such a dataset and published the relevant data and related source code. We hope this study to motivate the development and implementation of cost-sensitive learning techniques for healthcare problems in which the notion of data collection costs, human subject compliance, and patient discomfort are of paramount importance.

\section{Conclusion}
In many machine learning applications, especially in healthcare, it is of essential to consider feature acquisition costs. In this paper, we prepared a dataset consisting of nutrition and health data as well as feature acquisition costs for cost-sensitive learning in health domain. The prepared dataset consists of about 10,000 unique variables and can be used in defining various cost-sensitive studies in health. Furthermore, we compared the performance of the state-of-the-art approaches in the literature in three different classification problems including diabetes, heart disease, and hypertension classification.


\end{document}